# Person Retrieval in Surveillance Video using Height, Color and Gender

Hiren Galiyawala[1], Kenil Shah[2], Vandit Gajjar[1], Mehul S. Raval[1]
[1]School of Engineering and Applied Science, Ahmedabad University, India
[2]L. D. College of Engineering, India
{hiren.galiyawala, mehul.raval, vandit.gajjar}@ahduni.edu.in, shah.kenil.484@ldce.ac.in

**Abstract**

*A person is commonly described by attributes like height, build, cloth color, cloth type, and gender. Such attributes are known as soft biometrics. They bridge the semantic gap between human description and person retrieval in surveillance video. The paper proposes a deep learning-based linear filtering approach for person retrieval using height, cloth color, and gender. The proposed approach uses Mask R-CNN for pixel-wise person segmentation. It removes background clutter and provides precise boundary around the person. Color and gender models are fine-tuned using AlexNet and the algorithm is tested on SoftBioSearch dataset. It achieves good accuracy for person retrieval using the semantic query in challenging conditions.*

## 1. Introduction

Surveillance systems are deployed at many places for security. Commonly, person retrieval is done by manually searching through videos. Its primary goal is to localize the person(s) using semantic queries e.g., a tall male with a black t-shirt and blue jeans. Such descriptions are easy to understand and commonly used by an eye-witness to describe the person. However, such description cannot be fed to an automatic person retrieval system and they should be converted to a compatible representation.

The task of person retrieval in the video is very challenging due to occlusion, light condition, camera quality, pose, and zoom. However, attributes like height, cloth color, gender can be deduced from low-quality surveillance video at a distance without cooperation from the subject. Such attributes are known as soft biometrics [1, 21]. A single soft biometric cannot uniquely identify the individual. Therefore, it is important to find the most discriminative soft biometrics for person retrieval. A study identifies 13 commonly used soft biometric attributes [2].

Person retrieval using semantic description has been widely studied in recent years. Jain et al. [3] propose the integration of ethnicity, gender and height to improve performance of the traditional biometric system. Park et al. [4] develops visual search engine using dominant color, build and height to find the person. Techniques in [5 – 8] use soft biometrics for person re-identification which aims at locating a person of interest in camera network. The appearance-based model proposal by Farenzena et al. [5] uses three complementary aspects of the human appearance; the overall chromatic content, the spatial arrangement of colors into stable regions, and the presence of recurrent textures. Bazzani et al. [6] suggest an appearance-based model that incorporates a global and local statistical feature in the form of an HSV histogram and epitomic analysis respectively. However, techniques [5 – 8] are unsuitable for automatic identification and retrieval.

Description based person retrieval [9 – 11] uses an avatar generated from a semantic query. The avatar incorporates height and cloth color and search the frame using particle filter [9]. Denman et al. [10] generate a search query in the form of channel representation using height, dominant color (torso and leg), and clothing type (torso and leg).

Convolutional Neural Network (CNN) based approaches [12 – 14] are becoming popular for person attribute recognition. Multi-label convolutional neural network (MLCNN) [12] predicts gender, age, and clothing together with pedestrian attributes. Person's full body image is split into 15 overlapping 32×32 sized parts which are filtered and combined in the cost layer. Dangwei Li et al. [14] studies the limitation of handcrafted features (e.g., color histograms) and focuses on the relationship between different attributes. They propose deep learning based single attribute recognition model (DeepSAR) to independently identify the attribute. Another model (DeepMAR) exploits the relationship between the attributes to jointly recognize multiple attributes.

The approaches based on avatar [9, 10], channel representation [11] and CNN [12 – 14] considers bounding box around the person as potential area. This creates the following problems:

1. The bounding box with cluttered background may impact the person retrieval accuracy specifically for low-resolution video, occlusion, and multiple views.

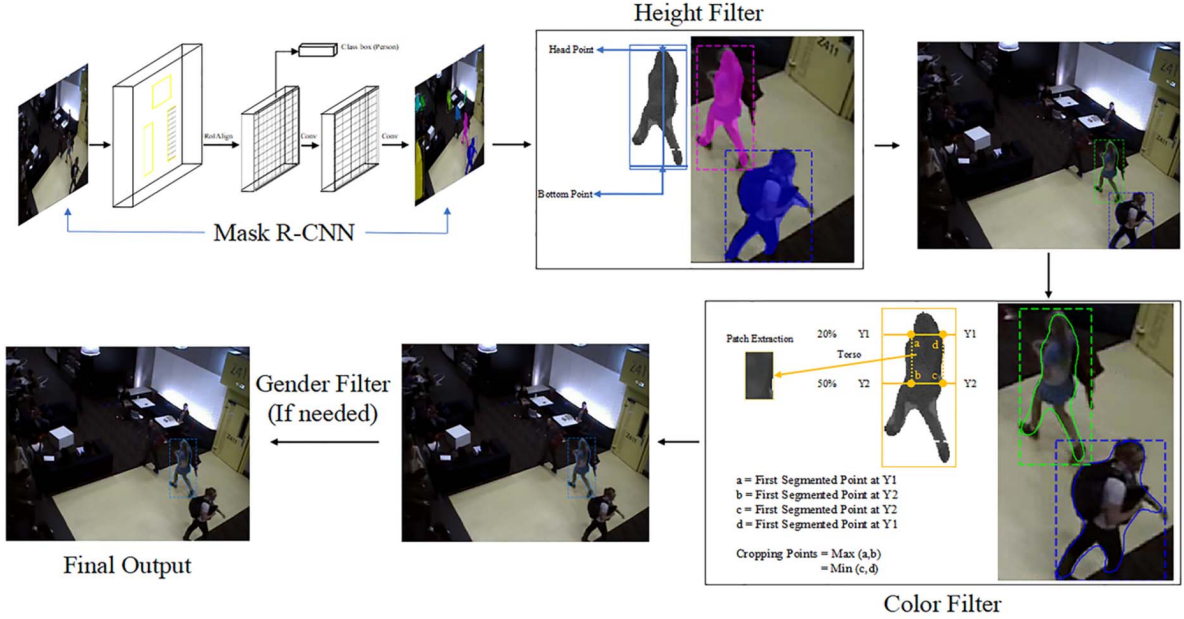

Figure 1: Proposed approach of person retrieval using height, cloth color and gender.

2. Person's height is estimated based on the bounding box. But, it may be larger in size than the person in the frame. This leads to incorrect height estimation and it also clutters the background.

This paper proposes a deep learning-based approach. It uses height, cloth color, and gender as linear filters for person retrieval. Such descriptors are chosen due to their discriminative ability and they also commonly occur in the investigation report. For example, height is view and distance invariant. As discussed in [15], the color prediction is also invariant to direction and view angle.

The height of the person is estimated using given camera calibration parameters while cloth color and gender are detected using CNN. Main contributions of this work are:

1. Use of semantic segmentation (pixel level segmentation) [16] to detect a person which has the following advantages:
   a. It removes unwanted background clutter.
   b. The precision of the segmented boundary yields accurate head and feet points. This results in a better estimate of the real-world height.
   c. It extracts precise patch for torso color classification.
2. The height can also be used to discriminate between the upright and sitting position of the person. This reduces search space for the person of interest in the upright position.
3. The paper proposes a generalized framework to deal with low resolution, view and pose variance.

The remainder of the paper is organized as follows. Section 2 describes models used for training data and fusion of various soft biometrics attributes to localize the person. Section 3 discusses the test results and the accuracy of the approach. Section 4 concludes the paper and discusses future work.

## 2. Proposed approach

This section introduces person retrieval based on height, cloth color, and gender. Figure 1 illustrates a flow diagram of the proposed framework. Each frame of video is given to state-of-the-art Mask R-CNN [16] for detection and pixel-wise segmentation of person(s). Head and feet points are extracted for all segmented persons. Using camera calibration parameters their respective height is computed which is compared with height given in the semantic query. Thus, height acts as a filter to reduce the number of persons in the frame. In case of multiple matches, further filtering is done using torso color. The use of semantic segmentation allows background free extraction of the color patch from the torso. The number of subjects is further reduced by matching the semantic query color with the classified color of extracted patches. The exactness of the final output is improved by using gender classification. Next subsections describe the process of height, color and gender estimation for person retrieval.

**Height estimation:** Person height is view-invariant and it also helps to discriminate between upright and sitting position of the person. Height is estimated using Tsai camera calibration approach [20], which translates the image coordinate to real-world coordinate.

SoftBioSearch dataset [10] provides 6 calibrated cameras for calculation of real-world coordinates. Person's head and feet points are estimated from the semantic segmentation (Figure 1). Steps for height estimation are as follows:

1. Intrinsic parameters matrix ($k$), rotation matrix ($R$) and a translation vector ($t$) are calculated from given calibration parameters.
2. The perspective transformation matrix is calculated as $C = k[R|t]$.
3. Head and feet points are undistorted using radial distortion parameters.
4. World coordinate for feet is set as $Z = 0$ $and$ $X, Y$ world coordinates are derived using inverse transformation of $C$.
5. Use $X, Y$ coordinate (of step-4) to calculate $Z$ coordinate of the head which also represent height.

Estimated height helps in reducing the search space within the test frame based on the description (e.g., 150 – 170 cm). Test frame would now contain only the person(s) which matches height description.

During training, annotated head and feet points are used to compute height from all frames of the video sequence. The height is then averaged ($H_{avg}$) over all frames. Over the same training sequence, it was observed that average height computed from automated head and feet point is larger than $H_{avg}$. This difference yields the wrong estimation, therefore; it is subtracted from average estimated height during testing to normalize the error.

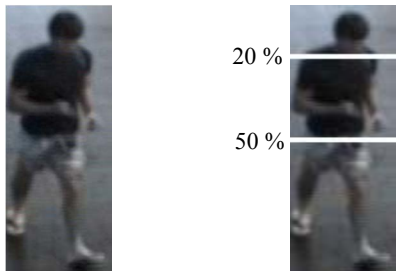

Figure 2: Extraction of torso and leg region from body.

**Torso color prediction:** Mask R-CNN generates a class label with probability score and semantic segmentation. The torso and leg regions are extracted using the golden ratio for height. The upper 20%-50% portion is classified as the torso, while the lower 50%-100% represents the legs of the person. The torso and leg segmentation are shown in Figure 2. Semantic segmentation is used to obtain the torso region without background to improve color classification. Color patch for prediction is extracted from region marked by points 'a', 'b', 'c', and 'd' (ref. Figure 1 – color filter). The extracted color patch is used to fine-tune AlexNet [18], to predict the probability score. The dataset [10] annotations contain two colors *Torso Color* and *Torso Second Color* for each subject. In case of multiple matches due to *Torso Color,* the algorithm will refine the result using *Torso Second Color* if present. This feature helps to further narrow down the search space.

**Gender classification:** The proposed algorithm accurately retrieves the person for most cases using height and cloth color. But, in the case of multiple matches, the algorithm uses gender classification. The AlexNet is fine-tuned using full body images for male and female gender classification.

## 2.1. Implementation details

This section covers details about the dataset, data augmentation, and AlexNet training. The proposed approach uses the weights of Mask R-CNN pretrained on Microsoft COCO [17] for detection and semantic segmentation. It has the Average Precision (AP) of 35.7 on COCO test set.

**Dataset overview:** This paper uses the SoftBioSearch database [10] which includes 110 unconstrained training video sequences, recorded using 6 stationary calibrated cameras. Each of the sequences contains 16 annotated soft biometric attributes and 9 body markers for subject localization. The 9 body markers are top of the head, left and right neck, left and right shoulder, left and right waist, approximate toe position of the feet. The test dataset contains video sequences of 41 subjects with the semantic query. The sequence length has 21 to 290 frames for training subjects. Discarding the frames with partial occlusion, the resulting training set has 8577 images from 110 subjects.

**Data augmentation:** It is a practice in deep learning to augment the training samples for improvement in performance and robustness. E.g., training AlexNet with 8577 images may result into over-fitting, which is avoided using data augmentation. Each training image is horizontally and vertically flipped, rotated with 10 angles {1º, 2º, 3º, 4º, 5º, -1º, -2º, -3º, -4º, -5º} and brightness increased with gamma factor of 1.5.

## 2.2. AlexNet training

The training is accomplished on a workstation with Intel Xeon core processor and accelerated by NVIDIA Quadro K5200 of 8 GB GPU. Color and gender models are fine-tuned using AlexNet which is pretrained on ImageNet [19] dataset.

**Color training:** The SoftBioSearch [10] database contains 1704 color patches divided into 12 culture colors. Additional patches for color training are extracted using 4 body markers namely left and right shoulder, left and right waist from the training dataset. In order to deal

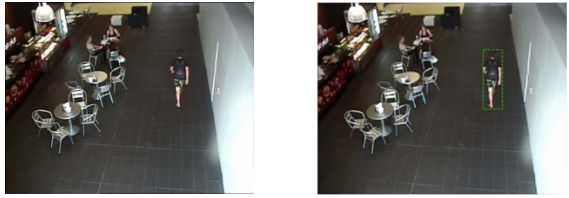

(a) Seq.10, F.59: height (170 – 190 cm), torso color (black) and gender (male). Person retrieved using only height.

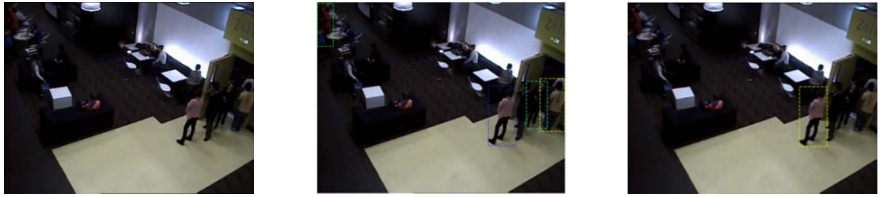

(b) Seq.04, F.76: height (160 – 180 cm), torso color (pink) and gender (female). Person retrieved using height and torso color.

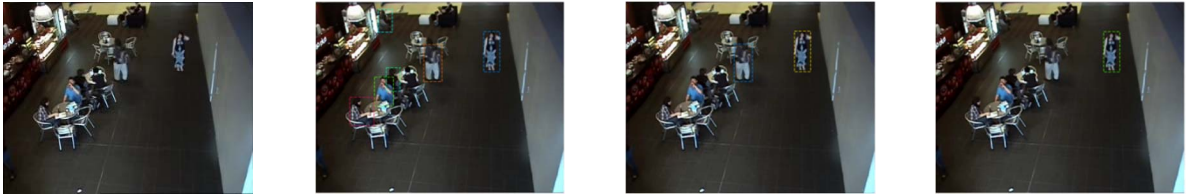

(c) Seq.8, F.31: height (130 – 160 cm), torso color (pink) and gender (female). Person retrieved using height, torso color and gender.
Figure 3: True positive cases of person retrieval with semantic query.

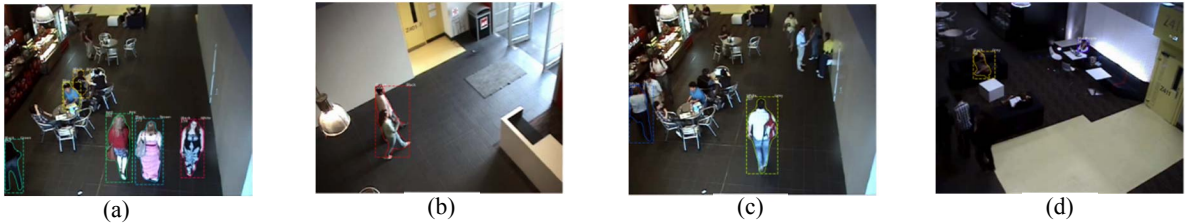

Figure 4: False negative cases of person retrieval. (a) incorrect color classification with multiple persons, (b) multiple persons with occlusion, (c) multiple persons with same torso color and height class and (d) person detection fails.

with illumination changes, these patches are augmented by increasing brightness with a gamma factor of 1.5. Approximately 17000 color patches are generated and further divided into 80% for training and 20% for the validation set. The last four layers of AlexNet namely (Conv5, fc6, fc7, and fc8) are fine-tuned for color training. It is trained for 30 epochs with a learning rate of 0.001, dropout set to 0.50 and effective batch size of 128. Above hyper-parameters led to an accuracy of 71.2% on the validation set.

**Gender training:** The data augmentation generated 105980 images for gender training which is approximately 13 times larger than the original training set (8577). The validation set has 20% of the total images. Gender training is accomplished by fine-tuning the last 3 layers (fc6, fc7, and fc8) of AlexNet. The model is fine-tuned for 20 epochs with learning rate 0.0001, batch size 64 images and a dropout rate of 0.40. This results in the accuracy of 68% on the validation set.

## 3. Experimental results and discussion

This section covers the qualitative and quantitative experimental results derived on test data set of 41 subjects. The ground truth (i.e., a person of interest) is established by manually mapping semantic test query to video frames of the test set. The ground truth is established after the first 30 initialization frames of each subject. Figure 3 shows true positive cases of the person retrieval using semantic queries. Abbreviations in the caption are interpreted as follows; e.g., Seq.10, F.59 indicates sequence number 10 with frame number 59 in the test set. In Figure 3, images from left to right indicate the input test frame, the output of the height filter, the output of color filter and gender filter respectively. Figure 3(a) shows a person of Seq.10, F.59 with semantic query height (170 – 190 cm), torso color (black) and gender (male). A person is retrieved using only a single soft biometric attribute i.e., height. Person of Seq.4, F.76 with semantic query height (160 – 180 cm), torso color (pink) and gender (female) is shown in Figure 3(b). It can be observed that multiple persons are detected with the same height class (ref. middle image in 3(b)). The correct person of interest is retrieved by adding torso color query to the height. Figure 3(c) shows a person of Seq.8, F.31 with semantic query height (130 – 160 cm), torso color (pink) and gender (female) in which algorithm utilizes all 3 semantic attributes i.e., height, torso color, and gender to retrieve person of interest. Thus, algorithm retrieves a person with a rank-1 match by utilizing a minimum number of attributes and in case of multiple matches, it uses additional soft biometrics to retrieve the person.

Figure 4 shows the results when the algorithm fails to retrieve the person correctly. It also indicates challenging conditions in the test dataset. Figure 4(a) shows a person of Seq.16, F.31 with semantic query height (160 – 180 cm), torso color (pink) and gender (female). It shows incorrect torso color classification with pink color classified as black (*Torso Color*) and brown (*Torso Second color*). It could be due to the presence of brown hair at the back side of the person. Figure 4(b) contains occlusion due to multiple persons. As a result, Mask R-CNN creates a single bounding box for Seq.25, F.34 with semantic query height (150 – 170 cm), torso color (green) and gender (female). Algorithm fails to retrieve person uniquely when multiple persons with the same height, torso color and gender are present, e.g., Seq.18, F.31 with semantic query height (180 – 210 cm), torso color (white) and gender (male) (Ref. Figure 4(c)). Figure 4(d) (Seq.1, F.73) shows the scenario where the person of interest is merged with a black background (except black and white torso region) due to poor illumination. Mask R-CNN fails to detect the person for such noisy scenarios. Person retrieval can be improved by incorporating additional attributes in the retrieval process. E.g., Figure 5 shows the improved result (Ref. Figure 4(c)) where the person is correctly retrieved using the leg color.

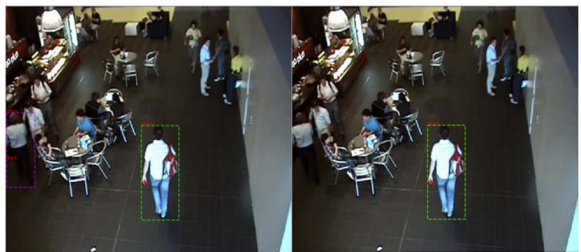

Figure 5: Use of leg color for improving retrieval.

In the proposed approach, each attribute is a filter which reduces the search space for person retrieval. The use of a linear filter for attribute classification has a potential weakness. The classification error due to first filter (height) will propagate to subsequent filters (color and gender).

True Positive (TP) rate is used as a qualitative measure of accuracy for a testing dataset of 41 persons. It is computed as follows:

$$TP(\%) = \frac{No.\ of\ frames\ with\ correct\ retrieval}{Total\ No.\ of\ frames} \quad (1)$$

The person localization accuracy for all frames is evaluated by computing an intersection-over-union (IoU) given by Eq. 2. Ground truth bounding box is constructed using the top of the head, the lowest of the feet, and the two most extreme locations from either feet, shoulder, waist or neck annotations.

$$IoU = \frac{D \cap GT}{D \cup GT} \quad (2)$$

where, D is the bounding box due to algorithm and GT is ground truth bounding box.

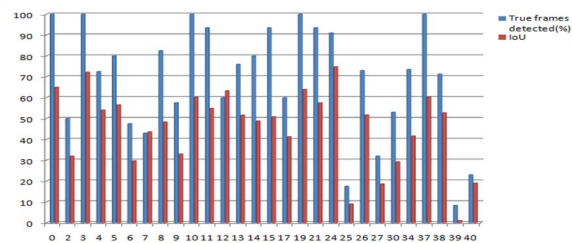

Figure 6: True Positive (TP) rate and IoU for the test set.

The algorithm correctly retrieves 28 out of 41 persons with varying TP rate. Figure 6 shows the TP rate (%) as well as IoU (Y-axis) and index of a correctly retrieved person (X-axis). Average TP rate of the algorithm is 65.8% and the resulting average IoU for correct retrieval is 0.458. Among 28 persons, 19 are retrieved with TP rate greater than 60%, 5 with TP rate between 30% - 60% and 4 persons with TP rate between 0% - 30%. For 19 persons (TP rate > 60%), height and color model works extremely well. The TP rate and IoU are very poor for some sequences. For example, in Seq.39 torso color (brown) is incorrectly classified which results in a TP rate of 8.32% and corresponding IoU as 0.101. This is due to a low number of color patches used in training and therefore, the model is unable to classify true color. The color classification can be improved by adding more color patches. In many frames, a person of interest is occluded e.g., Seq. 25 and semantic segmentation could not extract a precise boundary around the person. Similarly, in the Seq.1 person of interest is merged with background due to poor illumination resulting in poor detection.

## 4. Conclusion

The proposed approach retrieves the person in surveillance video using a semantic query based on height, cloth color, and gender. Use of semantic segmentation allows better height estimation and precise color patch extraction from the torso. The algorithm correctly recovers 28 persons out of 41 in a very challenging dataset with soft biometric attributes. The algorithm achieves an average IoU of 0.36 on the test dataset. It achieves an IoU of greater than or equal to 0.4 for 52.2% of frames. Future work will focus on improving results by incorporating other soft biometric attributes (e.g., leg color, torso texture, body accessory) and investigate mechanism to enhance robustness of the proposed approach.

## Acknowledgment

The authors would like to thank the Board of Research in Nuclear Sciences (BRNS) for a generous grant (36(3)/14/20/2016-BRNS/36020). We acknowledge the support of NVIDIA Corporation for a donation of the Quadro K5200 GPU used for this research. We would also like to thank the SAIVT Research Labs at Queensland University of Technology (QUT) for SAIVT-SoftBioSearch database.

## References

[1] A. Dantcheva, P. Elia, and A. Ross. What Else Does Your Biometric Data Reveal? A Survey on Soft Biometrics. IEEE Transactions on Information Forensics and Security 11(3): 441-467, 2015.

[2] M. D. MacLeod, J. N. Frowley, and J. W. Shepherd. Whole body information: Its relevance to eyewitnesses. In D. F. Ross, J. D. Read, & M. P. Toglia (Eds.), Adult eyewitness testimony: Current trends and developments, pp. 125 – 143. New York, US: Cambridge University Press 1994.

[3] A. K. Jain, S. C. Dass, and K. Nandakumar. Soft biometric traits for personal recognition systems. In International Conference on Biometric Authentication, Hong Kong, 2008, pp. 731–738.

[4] U. Park, A. Jain, I. Kitahara, K. Kogure, and N. Hagita. Vise: Visual search engine using multiple networked cameras. In 18th International Conference on Pattern Recognition (ICPR), 2006, pp. 1204 –1207.

[5] M. Farenzena, L. Bazzani, A. Perina, V. Murino, and M. Cristani. Person re-identification by symmetry-driven accumulation of local features. In IEEE Conference on Computer Vision and Pattern Recognition (CVPR), 2010, pp. 2360–2367.

[6] L. Bazzani, M. Cristani, A. Perina, M. Farenzena, and V. Murino. Multiple-shot person re-identification by HPE signature. In 20th International Conference on Pattern Recognition (ICPR), 2010, pp. 1413–1416.

[7] R. Zhao, W. Ouyang, and X. Wang. Person re-identification by salience matching. In IEEE International Conference on Computer Vision (ICCV), 2013, pp. 2528 – 2535.

[8] Y. Xu, L. Lin, W.-S. Zheng, and X. Liu. Human re-identification by matching compositional template with cluster sampling. In IEEE International Conference on Computer Vision (ICCV), 2013, pp. 3152 – 3159.

[9] S. Denman, M. Halstead, A. Bialkowski, C. Fookes, and S. Sridharan. Can you describe him for me? A technique for semantic person search in video. In International Conference on Digital Image Computing: Techniques and Applications, (DICTA), 2012, pp.1–8.

[10] M. Halstead, S. Denman, C. Fookes, and S. Sridharan, Locating people in video from semantic descriptions: a new database and approach. In International Conference on Pattern Recognition (ICPR), 2014 pp. 4501 – 4506.

[11] S. Denman, M. Halstead, C. Fookes, and S. Sridharan. Searching for people using semantic soft biometric descriptions. Pattern Recognition Letters, 68 (Part 2), pp. 306-315, 2015.

[12] J. Zhu, S. Liao, D. Yi, Z. Lei, and S. Z. Li. Multi-label CNN Based Pedestrian Attribute Learning for Soft Biometrics. In International Conference on Biometrics (ICB), 2015, pp. 535 – 540.

[13] P. Sudowe; H. Spitzer, and B. Leibe. Person Attribute Recognition with a Jointly-Trained Holistic CNN Model. In IEEE International Conference on Computer Vision Workshop (ICCVW), 2015, pp. 329 – 337.

[14] D. Li, X. Chen, and K. Huang. Multi-attribute Learning for Pedestrian Attribute Recognition in Surveillance Scenarios. In IAPR Asian Conference on Pattern Recognition (ACPR), 2015, pp. 111 – 115.

[15] P. Shah, M. S. Raval, S. Pandya, S. Chaudhary, A. Laddha, and H. Galiyawala. Description Based Person Identification: Use of Clothes Color and Type. In National Conference on Computer Vision, Pattern Recognition, Image Processing and Graphics (NCVPRIPG), Dec – 2017, IIT Mandi.

[16] K. He, G. Gkioxari, P. Dollár, and R. Girshick. Mask R-CNN. In IEEE International Conference on Computer Vision (ICCV), 2017, Oct 22, pp. 2980 – 2988.

[17] TY. Lin, M. Maire, S. Belongie, L. Bourdev, R. Girshick, J. Hays, P. Perona, D. Ramanan, C. L. Zitnick, and P. Dollár. Microsoft coco: Common objects in context. In European conference on computer vision, 2014, Sep 6, pp. 740 – 755.

[18] A. Krizhevsky, I. Sutskever, and G. E. Hinton. Imagenet classification with deep convolutional neural networks. In Advances in neural information processing systems, 2012 pp. 1097 – 1105.

[19] J. Deng, W. Dong, R. Socher, LJ. Li, K. Li, and L. Fei-Fei. Imagenet: A large-scale hierarchical image database. In IEEE conference on Computer Vision and Pattern Recognition (CVPR), 2009, Jun 20, pp. 248 – 255.

[20] R. Tsai, "A versatile camera calibration technique for high-accuracy 3D machine vision metrology using off-the-shelf TV cameras and lenses", IEEE Journal of Robotics and Automation, vol. 3, pp. 323-344, 1987.

[21] Mehul S Raval, "Digital Video Forensics: Description Based Person Identification", CSI Communications, Vol. 39, No. 12, pp. 9 - 11, March 2016.